%% file: main.tex
\documentclass{article}

\usepackage[preprint]{colm2026_conference}
\usepackage{fontspec}

\renewcommand{\encodingdefault}{T1}
\normalfont
\usepackage{microtype}
\usepackage{graphicx}
\usepackage{trimclip}
\usepackage{xcolor}
\usepackage{booktabs}
\usepackage{multirow}
\usepackage{array}
\usepackage{colortbl}
\usepackage{wrapfig}
\usepackage{placeins}
\usepackage{float}
\usepackage{amsmath}
\usepackage{algorithm}
\usepackage{algpseudocode}
\usepackage{tikz}
\usepackage{tcolorbox}
\usepackage{hyperref}
\usepackage{url}
\input{math_commands.tex}

\definecolor{abyss}{HTML}{121D36}
\definecolor{polarnight}{HTML}{1A2947}
\definecolor{nebula}{HTML}{2B3F66}
\definecolor{steeltrail}{HTML}{6D87BD}
\definecolor{skytrail}{HTML}{8FA8D8}
\definecolor{starlight}{HTML}{DFE7F5}
\definecolor{electricblue}{HTML}{3866FF}
\definecolor{covercream}{HTML}{EEF3FA}
\definecolor{coveraccent}{HTML}{3866FF}

\newfontfamily\outfit[
  Path=./,
  BoldFont=Outfit-SemiBold.ttf
]{Outfit-Regular.ttf}

\hypersetup{
  colorlinks=true,
  linkcolor=electricblue,
  citecolor=electricblue,
  urlcolor=coveraccent,
  filecolor=electricblue
}
\setcitestyle{numbers,square,comma,sort&compress}

\newcommand{\reporttitle}{PILOT in the Loop: Live Self-Improvement for Long-Horizon Agents}
\newcommand{\reportdisplaytitle}{\textcolor{electricblue}{PILOT} in the Loop: Live Self-Improvement for Long-Horizon Agents}
\title{\reporttitle}
\author{AllSpark Team}

\begin{document}
\raggedbottom
\fancyhead{}
\renewcommand{\headrulewidth}{0pt}
\color{abyss}
\thispagestyle{empty}

\vspace*{-0.44in}
\begin{tcolorbox}[
  width=\linewidth,
  colback=covercream,
  colframe=covercream,
  boxrule=0pt,
  arc=14pt,
  outer arc=14pt,
  boxsep=0pt,
  left=20pt,
  right=20pt,
  top=10pt,
  bottom=8pt
]
  {\outfit\fontsize{21.5}{25.5}\selectfont\bfseries\centering
    \reportdisplaytitle\par}
  \vspace{1.45em}
  {\bfseries\centering AllSpark Team\par}

  \vspace{0.75em}
  \begingroup
  \normalfont
  \setlength{\parindent}{0pt}
  \setlength{\parskip}{0pt}
  \input{sections/abstract}
  \par
  \endgroup

  \vspace{0.65em}
  \noindent
  \begin{minipage}[b]{0.63\linewidth}
    \outfit\fontsize{8.4}{10.2}\selectfont
    \textbf{Date:} August 25, 2026\\[-0.1em]
    \textbf{GitHub:} \href{https://github.com/XiaoYang66/Pilot}{github.com/XiaoYang66/Pilot}
  \end{minipage}%
  \hfill
  \begin{minipage}[b]{0.33\linewidth}
    \raggedleft
    \raisebox{-0.30em}{\includegraphics[height=16pt]{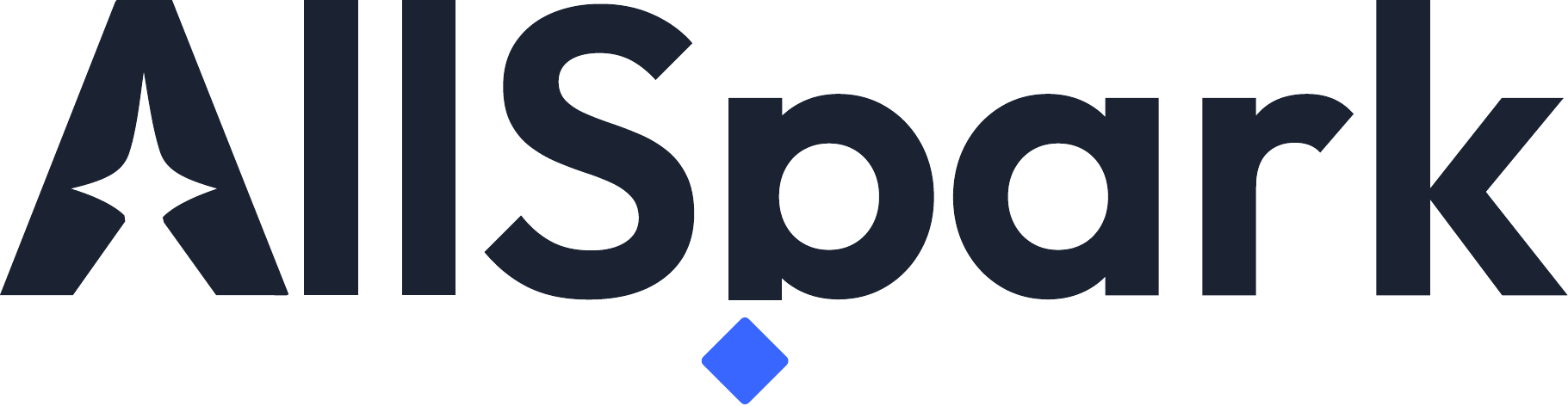}}%
  \end{minipage}
\end{tcolorbox}

\vspace{0.8em}
\begin{figure}[H]
\centering
\includegraphics[width=\linewidth]{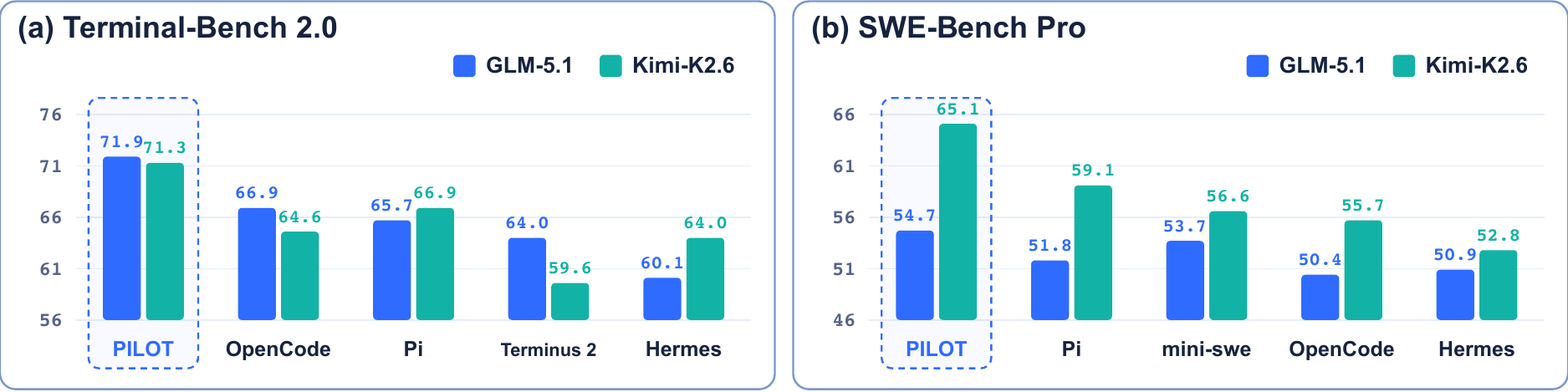}
\vspace{-0.65em}
\includegraphics[width=\linewidth]{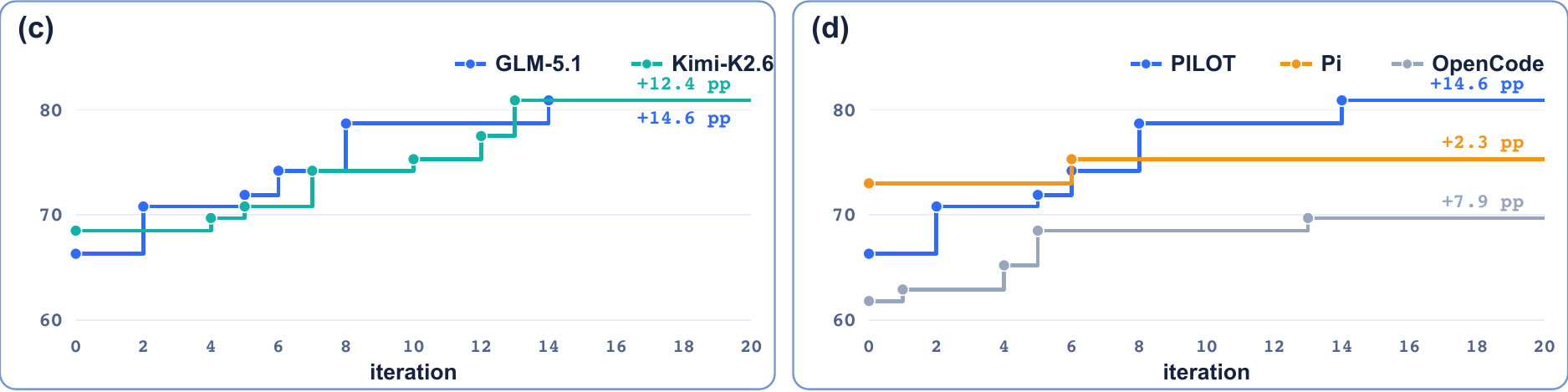}
\caption{Main results in two evaluation settings. (a)--(b) One-shot pass rate (\%) on Terminal-Bench~2.0 and SWE-bench Pro with GLM-5.1 and Kimi-K2.6; dashed boxes highlight \textsc{PILOT}. (c)--(d) Best-so-far Terminal-Bench~2.0 pass rate across 20 self-improvement iterations: (c) compares \textsc{PILOT} across backbones, and (d) compares \textsc{PILOT}, Pi, and OpenCode on GLM-5.1. Endpoint labels report improvement from iteration~0.}
\label{fig:iterations}
\end{figure}

\clearpage
\input{sections/report-body}

\bibliographystyle{colm2026_conference}
\bibliography{bibliography/references}

\appendix
\input{sections/appendix}

\end{document}

%% file: math_commands.tex
\usepackage{amsmath,amsfonts,bm}

\def\eqref#1{equation~\ref{#1}}

\def\1{\bm{1}}

\DeclareMathAlphabet{\mathsfit}{\encodingdefault}{\sfdefault}{m}{sl}
\SetMathAlphabet{\mathsfit}{bold}{\encodingdefault}{\sfdefault}{bx}{n}



%% file: sections/abstract.tex
Long-horizon agent runs generate experience that can improve both the current run and future runs: successful attempts reveal reusable procedures, while failed attempts expose failure modes.
Most self-improvement methods process this experience only after a run ends, so they can neither recover that run nor immediately apply and validate lessons learned from the run, making self-improvement less efficient and potentially less reliable.
We argue that self-improvement should instead be live, using emerging experience both to redirect the active run and to update the persistent harness.
However, live self-improvement exposes an architectural gap: existing agent architectures do not simultaneously support live correction and a dedicated self-improvement role.
Single-agent self-correction can revise the active run, but the same agent must execute the task and judge its trajectory within a limited context, splitting attention between execution and oversight.
Subagent delegation separates execution from the main agent, but the main agent typically cannot redirect the active subagent while the subagent is still running.
To bridge this architectural gap, we present \textbf{\textsc{PILOT}}, a supervisor--worker harness that realizes live self-improvement through two coupled mechanisms: (1)~\emph{live steering} lets a separate supervisor redirect or abort the active worker during execution; and (2)~\emph{live self-evolution} distils procedures and failure modes revealed during execution into reusable skills and memory.
Across two frozen backbones and three benchmarks, \textsc{PILOT} ranks first in five of six configurations.
On Terminal-Bench~2.0, \textsc{PILOT} outperforms counterpart harnesses by as much as 9.8 percentage points.
In the self-improvement setting, \textsc{PILOT} gains 14.6 points with GLM-5.1 and 12.4 points with Kimi-K2.6; mean output tokens fall by 42.9\% and 47.4\%, while successful evaluations per million output tokens rise by 110.3\% and 134.0\%, respectively.

%% file: sections/report-body.tex
\input{sections/intro}
\input{sections/method}
\input{sections/experiments}
\input{sections/casestudy}
\input{sections/related}
\input{sections/conclusion}

%% file: sections/intro.tex
\section{Introduction}

When agents work on long-horizon tasks, they generate experience that can improve both the active run and later work.
Successful trajectories reveal procedures worth retaining as reusable skills, particularly when a task can be solved but not yet solved reliably.
Failed or inefficient trajectories reveal strategies and recurring failure modes to avoid.
Agent self-improvement aims to learn from such experience and improve subsequent behavior.
A common approach is post-hoc review: reflection, judge-based evaluation, and self-evolving harness updates, all of which can improve the harness but cannot redirect the active run.
Reflection critiques a completed trajectory, judge-based evaluation assesses the final result, and self-evolving harnesses revise prompts, skills, or memory from completed traces and feedback \citep{reflexion,expel,autoharness,metaharness,ahe}.
These approaches can preserve experience, but their updates begin after the relevant execution has ended.
Newly extracted knowledge cannot help the run that produced it or be immediately applied and validated against the execution that revealed it.
Its usefulness can only be tested in a later rollout or a separate evaluation.
We therefore argue that self-improvement should be live, using emerging experience both to redirect the active run and to improve the persistent harness for later runs.

However, existing agent architectures do not simultaneously support live correction and a dedicated self-improvement role.
Single-agent self-correction is timely, but the same agent performs the task and diagnoses its own mistakes in a context filled with execution details \citep{react,selfrefine,critic}.
Execution details occupy the same context needed for diagnosis, making it harder for the agent to recognize and correct a strategy that is not working.
Subagent delegation separates execution from the main agent, but the main agent commonly receives the subagent's final summary only after the delegated run returns \citep{magenticone,claudecode-subagents,diveclaudecode}.
Subagent delegation therefore cannot usually redirect the active subagent while the subagent is still running.
Together, these limitations call for self-improvement that is both live and separate from task execution.

These requirements motivate \emph{live self-improvement}: a closed loop that uses emerging experience to redirect the active run and improve the persistent harness.
We present \textbf{\textsc{PILOT}}, a supervisor--worker harness that realizes live self-improvement through two coupled mechanisms: (1) \emph{live steering} lets the supervisor redirect the active worker during execution; and (2) \emph{live self-evolution} distils useful procedures, project conventions, and failure modes revealed during execution into reusable skills and memory.
\textsc{PILOT} makes self-improvement live by separating the roles of task execution and self-improvement, correcting the current run, using the live run to assess the correction, and retaining useful experience in the harness during execution.
In \textsc{PILOT}, a frozen \emph{supervisor} remains connected to a \emph{worker} throughout the worker's run rather than waiting for a completed trace.
Through a live channel, the supervisor receives the worker's questions and notifications together with the final result, execution errors, and inactivity alerts; the supervisor can then steer the worker's actions or abort the worker's run.
The worker's context absorbs execution details and dead ends, while the supervisor's separate context stays focused on the goal, recent events, and signs that the run is going off track.
The supervisor can also record reusable knowledge in the persistent harness without taking task execution away from the worker.

We evaluate \textsc{PILOT} in a one-shot setting and a self-improvement setting.
In the one-shot setting, \textsc{PILOT} ranks first in five of six backbone--benchmark combinations across two frozen backbones and three benchmarks.
On Terminal-Bench~2.0~\citep{terminalbench}, \textsc{PILOT} outperforms counterpart harnesses by as much as 9.8 percentage points.
In the self-improvement setting, we organize Terminal-Bench~2.0 tasks into iterations that share a fixed harness state.
Within each iteration, all tasks use that state, and agents receive no benchmark feedback during execution.
After the iteration, the harness retains skills distilled during successful runs for use in the next iteration.
\textsc{PILOT}'s best observed pass rate rises by 14.6 percentage points with GLM-5.1~\citep{glm-51} and 12.4 points with Kimi-K2.6~\citep{kimi-k26}.
Across iterations, the reusable skill set grows by 21 and 31 skills, mean output tokens per evaluated task fall by 42.9\% and 47.4\%, and successful evaluations per million output tokens rise by 110.3\% and 134.0\%, respectively.
Two case studies trace how the supervisor redirects an active strategy or corrects an implementation error while the worker remains responsible for the solution.

\noindent
We make three contributions: (1)~we formulate \emph{live self-improvement} as a closed loop that uses emerging experience both to redirect the active run and to update the persistent harness; (2)~we introduce \textsc{PILOT}, a supervisor--worker harness that couples live steering with live self-evolution over a shared stream of execution experience; and (3)~we evaluate \textsc{PILOT} across two frozen backbones and three long-horizon benchmarks, together with trajectory and efficiency analyses of task recovery, skill growth, and token efficiency.

%% file: sections/method.tex
\section{\textsc{PILOT}: A Live Self-Improvement Loop}
\label{sec:method}

\begin{figure}[t]
\centering
\includegraphics[width=\linewidth]{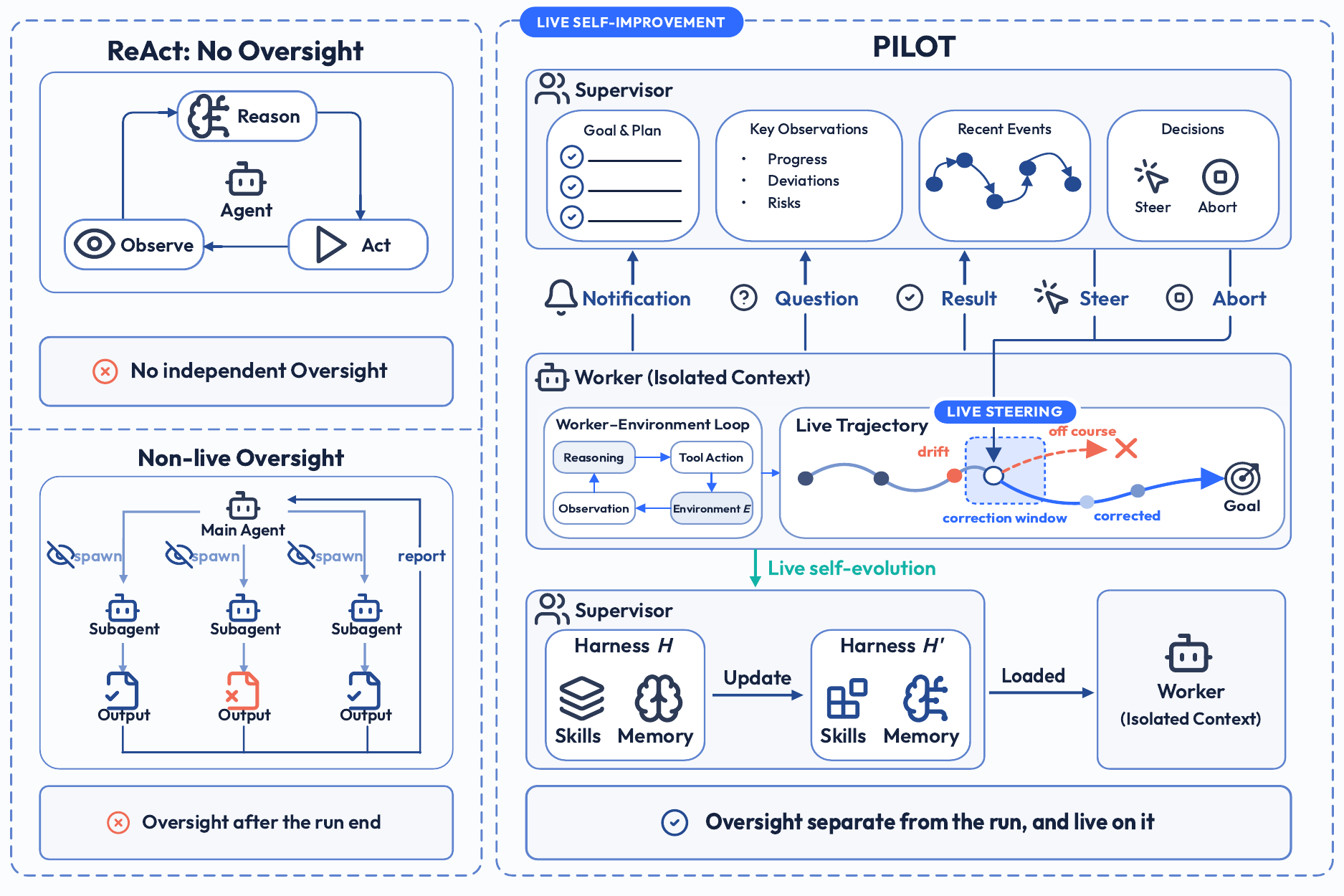}
\caption{\textsc{PILOT} implements live self-improvement through two mechanisms: live steering and live self-evolution. \emph{Top left}: a single ReAct loop has no oversight separate from the work. \emph{Bottom left}: a main agent receives a subagent's output only after the delegated run ends. \emph{Right}: a separate supervisor redirects or aborts the active worker, distils reusable skills and memory from supervision, and makes the evolved harness available to later workers.}
\label{fig:framework}
\end{figure}

\textsc{PILOT} treats live self-improvement as a closed loop rather than an update performed only after execution.
With the worker responsible for task execution and the supervisor responsible for self-improvement, the supervisor uses \emph{live steering} to redirect the active worker during execution and \emph{live self-evolution} to record reusable knowledge in the persistent harness (Figure~\ref{fig:framework}).
Worker sessions spawned after a harness update load the evolved harness and enter the same two loops.
Appendix~\ref{app:pilot-pseudocode} provides pseudocode for the two coupled loops.

\paragraph{Roles and persistent state.}
A long-horizon task $\tau$ is attempted in a single episode in an environment $\mathcal{E}$ whose state changes with the worker's actions.
The model parameters $\theta$ remain frozen, while a persistent harness $H$ includes a skill library $\mathcal{K}$ and memory $\mathcal{M}$ that persist across episodes.
Throughout this paper, \emph{self-improvement} therefore refers to the evolution of the persistent harness $H$, not an update to the model parameters $\theta$.
During an episode, the supervisor can spawn one or more workers, $W_j \gets \textsc{Spawn}(\theta, \tau_j, H)$, either concurrently or as earlier workers settle.
Each worker loads the current harness $H$, operates in an isolated context, and produces a trajectory $\xi_j$ of actions and observations in $\mathcal{E}$.
The workers' exploration, dead ends, and verbose tool output remain in their isolated contexts by default.
The supervisor reads the relevant portion of each $\xi_j$ only when diagnosis is needed, preserving the supervisor's context for the goal, recent events, and recurring failure patterns.

\paragraph{Live steering through a two-way channel.}
During the episode, the supervisor remains connected to each active worker through a two-way live channel.
For every worker session, the channel supports three worker-to-supervisor events and two supervisor-to-worker actions.
\textbf{(1) Notification.}
Each worker decides when to report progress, an intermediate result, or a potential risk; its execution continues after the notification is sent.
\textbf{(2) Question.}
A worker decides when its next step requires supervisory input and pauses until the supervisor replies.
\textbf{(3) Result.}
The runtime automatically delivers a worker's final result and worker index $j$ to the supervisor when that worker finishes.
\textbf{(4) Steer.}
When live evidence indicates that a worker's current course should change, the supervisor can inspect the relevant portion of $\xi_j$ and queue guidance for that worker's next turn; the current turn finishes first.
\textbf{(5) Abort.}
The supervisor interrupts an active worker $W_j$ when continuing that worker session is no longer useful.

\paragraph{Live self-evolution.}
An active run can reveal a successful procedure or project convention worth reusing, or a recurring failure mode worth avoiding.
When the supervisor identifies such knowledge in the live trajectory, the supervisor records it in $\mathcal{K}$ or $\mathcal{M}$, refining the harness from $H$ to $H'$.
Every worker spawned after the update, within the same episode or in a later episode, loads $H'$.
The new worker's trajectory is again exposed to live steering and live self-evolution, closing the self-improvement loop.

\paragraph{Implementation.}
\textsc{PILOT} is implemented as an extension to the Pi coding-agent runtime~\citep{zechner2026pi}; the supervisor is an agent session, and workers are spawned in-process as separate sessions.
Our experiments follow a real-world usage scenario in which the same frozen model serves as both supervisor and worker.

%% file: sections/experiments.tex
\section{Experiments}
\label{sec:experiments}
\FloatBarrier

\begin{table}[t]
\centering
\small
\caption{Pass rate (\%) across three long-horizon agent benchmarks, with the same frozen open backbone driving every harness. (a) Terminal-Bench~2.0 by difficulty; \emph{All} aggregates its 89 tasks and \emph{AVG} averages the two backbones' All rates. (b) SWE-bench Multilingual and SWE-bench Pro; \emph{AVG} averages the two backbones. Column maxima are bold.}
\label{tab:main-results}
\textbf{(a) Terminal-Bench~2.0}\par\vspace{0.35em}
\setlength{\tabcolsep}{5pt}
\begin{tabular*}{0.96\linewidth}{@{\extracolsep{\fill}} l *{9}{c}}
\toprule
\multirow{2}{*}{Harness} & \multicolumn{4}{c}{GLM-5.1} & \multicolumn{4}{c}{Kimi-K2.6} & \multirow{2}{*}{AVG} \\
\cmidrule(lr){2-5}\cmidrule(lr){6-9}
& Easy & Medium & Hard & All & Easy & Medium & Hard & All & \\
\midrule
Terminus-2 & 75.0 & 72.7 & 46.7 & 64.0 & 87.5 & 69.1 & 38.3 & 59.6 & 61.8 \\
Hermes & 87.5 & 67.3 & 43.3 & 60.1 & 87.5 & 77.3 & 36.7 & 64.0 & 62.1 \\
OpenCode & \textbf{100.0} & 75.5 & 46.7 & 66.9 & 75.0 & 75.5 & 43.3 & 64.6 & 65.8 \\
Pi & 87.5 & 72.7 & 50.0 & 65.7 & 100.0 & 74.5 & 48.3 & 66.9 & 66.3 \\
\midrule
\textbf{\textsc{PILOT}} (ours) & 87.5 & \textbf{80.0} & \textbf{55.0} & \textbf{71.9} & \textbf{100.0} & \textbf{78.2} & \textbf{55.0} & \textbf{71.3} & \textbf{71.6} \\
\bottomrule
\end{tabular*}
\par\vspace{0.85em}
\textbf{(b) Software-engineering benchmarks}\par\vspace{0.35em}
\begin{tabular*}{0.96\linewidth}{@{\extracolsep{\fill}} l *{6}{c}}
\toprule
\multirow{2}{*}{Harness} & \multicolumn{3}{c}{SWE-bench Multilingual} & \multicolumn{3}{c}{SWE-bench Pro} \\
\cmidrule(lr){2-4}\cmidrule(lr){5-7}
& GLM-5.1 & Kimi-K2.6 & AVG & GLM-5.1 & Kimi-K2.6 & AVG \\
\midrule
Hermes         & 67.5 & 72.2 & 69.9 & 50.9 & 52.8 & 51.9 \\
OpenCode       & 68.9 & 74.1 & 71.5 & 50.4 & 55.7 & 53.1 \\
Mini-SWE-Agent & 68.5 & 73.7 & 71.1 & 53.7 & 56.6 & 55.2 \\
Pi             & 69.8 & \textbf{75.9} & \textbf{72.9} & 51.8 & 59.1 & 55.5 \\
\midrule
\textbf{\textsc{PILOT}} (ours) & \textbf{71.6} & 73.7 & 72.7 & \textbf{54.7} & \textbf{65.1} & \textbf{59.9} \\
\bottomrule
\end{tabular*}
\end{table}

\subsection{Setup}
\label{sec:experiment-setup}

\textbf{Benchmarks.}
We evaluate on three benchmarks of long-horizon, real-world agent tasks: (1) Terminal-Bench~2.0, whose 89 runnable tasks require an agent to drive a shell over many steps to reach a system or engineering goal; (2) SWE-bench Multilingual \citep{swebenchmm}, which extends repository-level software repair beyond Python; and (3) SWE-bench Pro \citep{swebenchpro}, which contains longer and harder repository-level issues than the original SWE-bench \citep{swebench}.
Terminal-Bench~2.0, SWE-bench Multilingual, and SWE-bench Pro together span terminal operation and multi-language code repair, two demanding forms of long-horizon agent work.

\textbf{Backbones.}
We use two open-weights models as frozen backbones, Kimi-K2.6 and GLM-5.1.
Within each condition, the same frozen backbone fills both the supervisor and the worker.
Using the same frozen model in both roles isolates the supervisor--worker orchestration from any capability gap between the supervisor and the worker (\S\ref{sec:method}).

\textbf{Harness baselines.}
On Terminal-Bench~2.0 we compare \textsc{PILOT} against four single-agent harnesses driven by the same backbone: Pi, the coding agent that \textsc{PILOT} extends; OpenCode~\citep{opencode}; Terminus-2~\citep{terminus2}, maintained by the Terminal-Bench team; and Hermes~\citep{hermesagent}.
On SWE-bench Multilingual and SWE-bench Pro we compare against Pi, OpenCode, Mini-SWE-Agent~\citep{minisweagent}, and Hermes on the same backbones.
Holding the backbone fixed across harnesses isolates the effect of the harness from the effect of the model.

\textbf{Evaluation settings.}
We evaluate the two parts of \textsc{PILOT}'s live self-improvement loop in settings that reflect real-world agent use.
(1) \emph{One-shot setting.}
A developer hands the harness a new task whose environment changes as work proceeds.
Every task starts from a fresh harness state, isolating whether live steering keeps the worker on track during the run.
(2) \emph{Self-improvement setting.}
A developer or team returns to the same harness for related work.
To evaluate the complete loop, including whether live self-evolution turns experience accumulated during supervision into reusable harness knowledge, we organize the Terminal-Bench~2.0 runs into iterations, where one iteration is a complete sweep over the benchmark tasks.
At the start of iteration $i$, every run receives an isolated copy of the same shared harness state $H_i$, which includes the skill library $\mathcal{K}_i$ and memory $\mathcal{M}_i$.
While a task is running, the harness can create or revise skills and memory using only the live agent trajectory and environment feedback.
These updates occur during task execution, before the verifier outcome is available; neither the supervisor nor the worker receives any benchmark evaluation signal or reward.
Thus, a candidate update is skill or memory content created or revised during a run, not a post-hoc summary generated from its evaluation result.
After every run in iteration $i$ has finished, verifier outcomes only decide which updates are carried into $H_{i+1}$: updates from successful runs are retained, whereas updates from failed runs are not.
Verifier outcomes are never used to create or modify those updates.
Every run in iteration $i+1$ then starts from the merged shared state $H_{i+1}$.
For the backbone comparison, \textsc{PILOT} runs with GLM-5.1 and Kimi-K2.6 begin from the same $H_0$; for the harness comparison, the GLM-5.1 runs of \textsc{PILOT}, Pi, and OpenCode also begin from $H_0$, with each configuration updating independently thereafter.
We measure skill-library size as the number of distinct skills in $\mathcal{K}_i$.
This design mirrors a real-world use scenario in which developers explicitly instruct an agent harness to retain reusable skills across related tasks.
Accordingly, the setting includes a dedicated self-improvement instruction that specifies how to reuse and record experience.
Every evaluated harness receives the same instruction throughout the experiment; Appendix~\ref{app:self-improvement-instruction} provides the full text.

\textbf{Protocol.}
Each backbone runs with the enable-thinking and preserve-thinking settings on, a maximum generation length of 32k tokens, and all other parameters left at the official defaults for that backbone.
We run each configuration twice and report the mean pass rate across the two runs.
Every task runs in an isolated container sandbox.
On Terminal-Bench~2.0, following common practice, each task is capped at three hours of wall-clock time, with CPU, memory, and other resources left at the task's own configuration.
On SWE-bench Multilingual and SWE-bench Pro, a few tasks pin environment versions that conflict with the environment some harnesses require; we apply the same exclusions to every harness. Appendix~\ref{app:excluded-swe} describes the exclusion criterion.
For each iteration, we compute the mean output tokens generated per evaluated task; for \textsc{PILOT}, the calculation includes all supervisor and worker assistant turns.
Figure~\ref{fig:distillation-analysis}(c) then averages these per-iteration means within each five-iteration window.

\subsection{Live steering keeps long-horizon work on track}
\label{sec:live-steering-results}

We evaluate \textsc{PILOT} in the one-shot setting while holding the backbone fixed across all harnesses.
We first compare the harnesses on Terminal-Bench~2.0 and then analyze generalization across backbone models and task domains.

\paragraph{Live steering improves performance on long-horizon tasks.}
As shown in Table~\ref{tab:main-results}(a), \textsc{PILOT} reaches the highest one-shot pass rate on Terminal-Bench~2.0 with both frozen backbones.
\textsc{PILOT} reaches 71.9\% on GLM-5.1, 5.0 percentage points above OpenCode at 66.9\%, and 71.3\% on Kimi-K2.6, 4.4 points above Pi at 66.9\%.
Across the two backbones, \textsc{PILOT} averages 71.6\%, 5.3 points above Pi at 66.3\%, the strongest single-agent baseline by average.
On hard tasks, \textsc{PILOT} reaches 55.0\% on each backbone, exceeding Pi by 5.0 points on GLM-5.1 (50.0\%) and 6.7 points on Kimi-K2.6 (48.3\%).

\paragraph{Generalization across backbone models and task domains.}
As shown in Table~\ref{tab:main-results}, \textsc{PILOT}'s live-steering mechanism generalizes across both backbone models and task domains.
Across backbone models, \textsc{PILOT} ranks first on all three benchmarks with GLM-5.1; with Kimi-K2.6, \textsc{PILOT} ranks first on Terminal-Bench~2.0 and SWE-bench Pro and second on SWE-bench Multilingual.
The results also extend across task domains.
Terminal-Bench~2.0 evaluates interactive terminal work, while SWE-bench Pro evaluates hard repository-level software repair.
Pi, a widely used single-agent harness, is the strongest average baseline on both benchmarks.
On Terminal-Bench~2.0, \textsc{PILOT} averages 71.6\%, 5.3 points above Pi at 66.3\%.
On SWE-bench Pro, \textsc{PILOT} averages 59.9\%, 4.4 points above Pi at 55.5\%.
Overall, \textsc{PILOT} ranks first in five of the six combinations, showing that the gains are not confined to a particular backbone model or task domain.
\FloatBarrier

\subsection{Live self-evolution makes supervision reusable across tasks}
\label{sec:live-evolution-results}

Real-world harness use often spans a sequence of related long-horizon tasks.
We evaluate whether live self-evolution closes the self-improvement loop by turning supervision from earlier tasks into reusable knowledge for later tasks.
We first measure the complete \textsc{PILOT} loop across backbone models and then compare \textsc{PILOT} with Pi and OpenCode, all initialized from the same skill library.

\paragraph{The closed loop yields consistent gains across iterations and backbones.}
As shown in Figure~\ref{fig:iterations}(c), \textsc{PILOT}'s best observed pass rate on GLM-5.1 rises from 66.3\% at iteration~0 to 80.9\%, a gain of 14.6 percentage points.
On Kimi-K2.6, \textsc{PILOT}'s best observed pass rate rises from 68.5\% to 80.9\%, a gain of 12.4 points.
The harness accumulates reusable skills and memory across iterations.
The consistent gains on GLM-5.1 and Kimi-K2.6 show that the benefit of closing the loop between supervision and persistent harness updates is not specific to one backbone.

\paragraph{Live self-evolution strengthens both skill accumulation and reuse.}
Figure~\ref{fig:iterations}(d) compares \textsc{PILOT}, Pi, and OpenCode using the same frozen GLM-5.1 backbone and the same initial skill library.
For every task, all three harnesses receive identical user input, including the same task description and instructions.
After iteration~0, each harness incorporates experience from its own runs into its skill library.
Figure~\ref{fig:iterations}(d) therefore measures how effectively each harness turns accumulated experience into performance on later tasks.
From iteration~0 to each harness's best observed result, \textsc{PILOT} improves by 14.6 points, compared with 7.9 points for OpenCode and 2.3 points for Pi.
The stronger gains over Pi and OpenCode show that, in the real-world self-improvement scenario, \textsc{PILOT} both accumulates more useful experience and applies that experience more effectively to later tasks.

%% file: sections/casestudy.tex
\input{sections/figure-self-evolution-analysis}

\section{Analysis of Live Self-Improvement}
\label{sec:analysis}

Section~3 shows that \textsc{PILOT} improves the active run through live steering and the persistent harness through live self-evolution.
To understand how live self-improvement produces these gains, we first examine task coverage, skill accumulation, and token efficiency under live self-evolution, then analyze which successful runs are aided by live steering and illustrate two representative corrections.

\setcounter{topnumber}{1}

\subsection{Live self-evolution makes accumulated experience more useful}
\label{sec:repeated-analysis}

Section~\ref{sec:live-evolution-results} shows that the complete \textsc{PILOT} loop improves Terminal-Bench~2.0 performance across iterations.
We now test whether those gains coincide with the persistent changes expected from live self-evolution: broader task coverage, a growing skill set, and more efficient execution.

\textbf{Performance gains extend across all three difficulty levels and are largest on Hard tasks.}
GLM-5.1 first reaches its maximum at iteration~14, while Kimi-K2.6 first reaches its maximum at iteration~13.
Figure~\ref{fig:distillation-analysis}(a) shows the corresponding gain for each difficulty level.
GLM-5.1 gains 2 additional passes on Easy, 6 on Medium, and 8 on Hard; Kimi-K2.6 gains 1, 7, and 12 additional passes, respectively.
The largest gains occur on Hard tasks.
Hard tasks more often require specific procedures, recovery strategies, and tool-use patterns that the backbone does not reliably reconstruct from scratch.
By distilling these patterns into reusable harness knowledge, \textsc{PILOT} reduces repeated exploration and makes later attempts more reliable.

\textbf{Live self-evolution increases the number of reusable skills.}
As shown in Figure~\ref{fig:distillation-analysis}(b), the number of GLM-5.1 skills grows from 62 to 83, while the number of Kimi-K2.6 skills grows from 50 to 81.
The consistent growth across both backbones shows that \textsc{PILOT} does not treat each run as an isolated task; it turns execution experience into an expanding pool of reusable procedures.

\textbf{The closed loop improves token efficiency across iterations.}
Figure~\ref{fig:distillation-analysis}(c) reports mean output tokens per evaluated task, counting all supervisor and worker assistant turns and averaging the per-iteration means across five-iteration windows.
The mean falls from 28.5K to 16.3K tokens per evaluated task on GLM-5.1, a 42.9\% reduction, and from 41.9K to 22.1K on Kimi-K2.6, a 47.4\% reduction.
Figure~\ref{fig:distillation-analysis}(d) reports successful evaluations per million output tokens.
The best observed efficiency rises by 110.3\% on GLM-5.1 and 134.0\% on Kimi-K2.6 relative to iteration~0.
Together, the lower generation cost and higher success per token show that accumulated harness knowledge reduces repeated reasoning and exploration, allowing later tasks to better reuse established procedures.

\subsection{Live steering concentrates on harder tasks}

\begin{wraptable}{r}{0.48\linewidth}
\vspace{-1.5em}
\centering
\small
\caption{Live-steering analysis in the one-shot setting: percentage of successful Terminal-Bench~2.0 runs classified as aided by live steering, grouped by task difficulty and backbone.}
\label{tab:tb2-supervision}
\begin{tabular}{lrr}
\toprule
Difficulty & GLM-5.1 (\%) & Kimi-K2.6 (\%) \\
\midrule
Easy   & 0.0 & 0.0 \\
Medium & 1.1 & 8.1 \\
Hard   & 6.1 & 19.7 \\
All    & 2.3 & 10.6 \\
\bottomrule
\end{tabular}
\vspace{-0.6em}
\end{wraptable}

Section~\ref{sec:live-steering-results} shows that live steering improves one-shot performance on long-horizon tasks.
Within the one-shot setting, we manually inspect the complete supervisor--worker trajectories to identify successful Terminal-Bench~2.0 runs aided by live steering.
We classify a successful run as aided by live steering only when the trace shows that the supervisor identifies a concrete error, stalled branch, or unproductive strategy and provides corrective direction; the worker or a replacement branch follows that direction; and the worker ultimately completes the task successfully along the corrected path.
An intervention followed by PASS is not sufficient: we exclude interventions that are ignored, stale, or redundant, as well as cases in which success follows an unrelated execution path.
Table~\ref{tab:tb2-supervision} reports the percentage of all successful runs classified as aided by live steering.
No successful Easy run is classified as aided by live steering.
From Medium to Hard, the share rises from 1.1\% to 6.1\% for GLM-5.1 and from 8.1\% to 19.7\% for Kimi-K2.6; the overall shares are 2.3\% and 10.6\%, respectively.
Trace-supported correction contributions are absent from the Easy split and are more frequent on Hard than Medium tasks for both backbones.
Easy tasks generally fall within the worker's existing capabilities and can be completed without external redirection.
Harder tasks require longer, more fragile execution chains in which errors can compound, leaving more opportunities for the supervisor to recover the run.
The pattern suggests that live steering is most useful for difficult tasks whose long execution horizon creates greater risk of drift or stalled progress.

\paragraph{Representative live-steering cases.}
Figure~\ref{fig:live-steering-cases} illustrates two successful trajectories in which the supervisor identifies a concrete problem and the worker follows the resulting correction.
The first case redirects an unproductive strategy, while the second corrects an implementation error.

\textsc{PILOT} makes these mid-run corrections possible by separating execution from oversight.
The worker keeps its context focused on tool use, intermediate results, and implementation, while the supervisor maintains an outside view of the goal, recent events, and deviations from the plan.
This division of responsibility keeps each agent's attention focused on its own role and allows the supervisor to detect a stalled or incorrect branch before the run is lost.
The worker remains responsible for the final solution; the supervisor provides timely direction that helps the worker recover.

\input{sections/figure-live-steering-cases}
\FloatBarrier

%% file: sections/figure-self-evolution-analysis.tex
\begin{figure}[t]
\centering
\includegraphics[width=\linewidth]{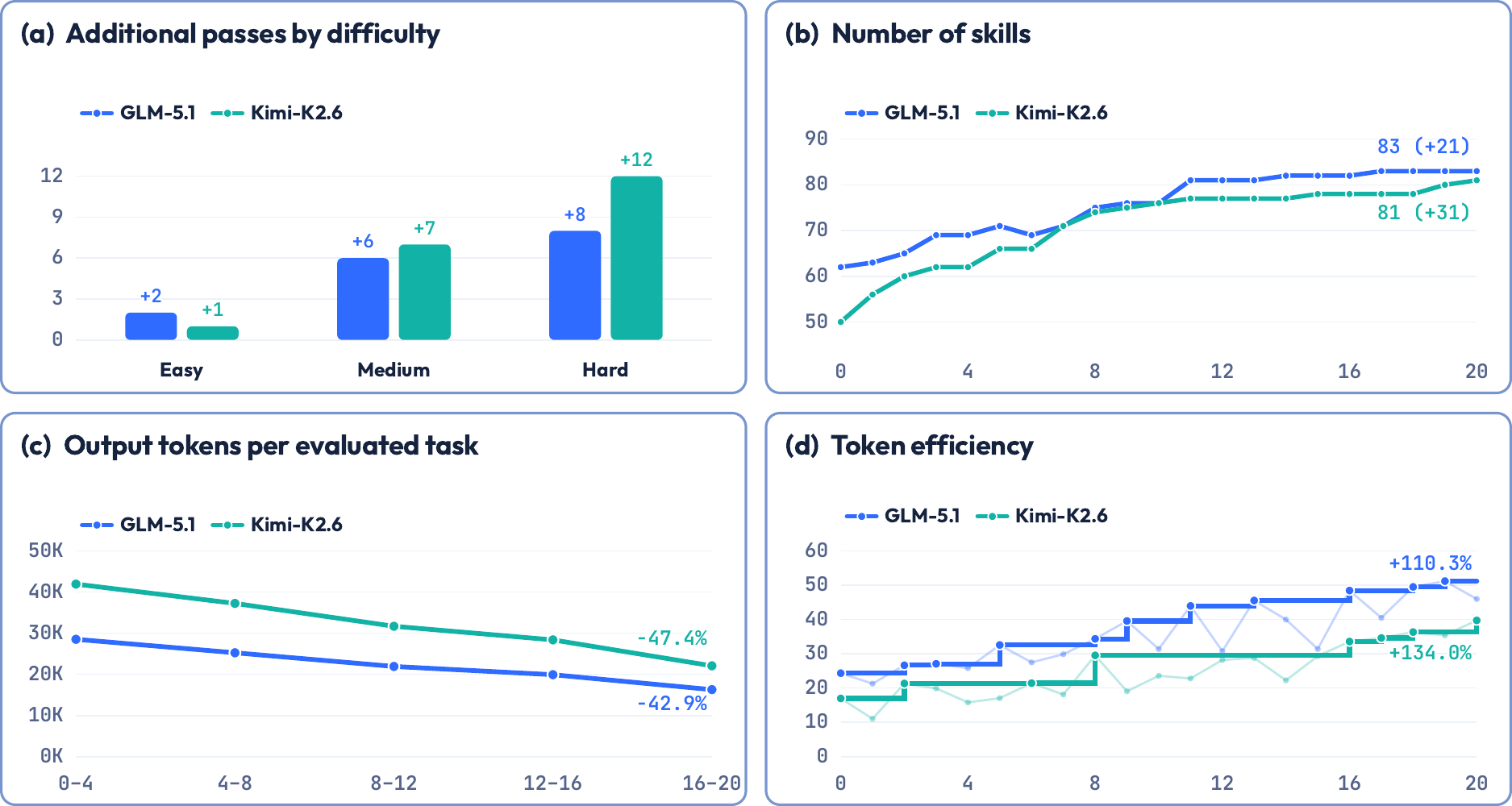}
\caption{\textsc{PILOT} across iterations in the self-improvement setting. (a) Additional passing tasks from iteration~0 to the first maximum for each backbone, separated by task difficulty. (b) Number of distinct skills available to \textsc{PILOT} at each iteration. (c) Mean output tokens per evaluated task, averaged over iteration ranges 0--4, 4--8, 8--12, 12--16, and 16--20; \textsc{PILOT} includes all supervisor and worker assistant turns. (d) Successful evaluations per million output tokens. Thin lines in (d) show the value at each iteration, thick step lines show the best value observed at or before that iteration, and endpoint labels report improvement relative to iteration~0.}
\label{fig:distillation-analysis}
\end{figure}

%% file: sections/figure-live-steering-cases.tex
\begin{figure}[t]
\centering
\begingroup
\setlength{\fboxsep}{7pt}
\noindent\fbox{%
\begin{minipage}{0.95\linewidth}
\small
\textbf{(a) Strategy-level correction: \texttt{winning-avg-corewars} (Medium), Kimi-K2.6}

\smallskip
\textbf{Worker trajectory before live steering.}
The worker spent over twenty minutes tuning DAT-clear variants.
The experiments repeatedly produced ties against multi-process opponents, and the required win rates remained unmet.

\smallskip
\textbf{Live steering by the supervisor.}
The supervisor sent the following live correction:
``Stop testing synthetic opponents and change strategy completely. \ldots{}
Find a published warrior source \ldots{} and test it against the actual five opponents.''

\smallskip
\textbf{Worker adoption.}
The worker immediately acknowledged the correction:
``The supervisor is right. \ldots{} I need to pivot to using a proven classic warrior.''

\smallskip
\textbf{Worker trajectory after live steering.}
The worker consulted public CoreWars resources, adapted Silk Warrior~1.3, and tuned its replication strategy against the five actual opponents.
The worker retrieved the source from the public \texttt{mbarbon/corewar-koth} archive; the trace contains no access to the Terminal-Bench repository or a benchmark solution.
The final warrior passed every threshold: Stone 93\%, Paper 94\%, Vampire 84\%, Snake 55\%, and G2-Clear 53\%.
\end{minipage}}
\endgroup
\par\medskip
\begingroup
\setlength{\fboxsep}{7pt}
\noindent\fbox{%
\begin{minipage}{0.95\linewidth}
\small
\textbf{(b) Implementation-level correction: \texttt{torch-tensor-parallelism} (Hard), GLM-5.1}

\smallskip
\textbf{Worker trajectory before live steering.}
The worker computed \texttt{RowParallelLinear} with
\texttt{F.linear(x, self.weight, self.bias)} and then summed the partial outputs with \texttt{all\_reduce}.
This added the full bias once per rank.

\smallskip
\textbf{Live steering by the supervisor.}
The supervisor sent the following live correction:
``Important correction for \texttt{RowParallelLinear}: The forward method should NOT include the bias parameter in \texttt{F.linear}.
\ldots{} The bias should be added after the \texttt{all\_reduce}, not before.''

\smallskip
\textbf{Worker adoption.}
The worker explicitly accepted the correction:
``The supervisor is correct.
\ldots{} the result after summing would be \texttt{world\_size * bias} instead of just \texttt{bias}.''
The worker then changed the code to \texttt{F.linear(x, self.weight, None)}, followed by \texttt{all\_reduce} and a single bias addition.

\smallskip
\textbf{Worker trajectory after live steering.}
A later repair replaced the bare communication calls with autograd-compatible operations while preserving the corrected bias placement.
The final implementation passed all 13 verifier tests across world sizes 1, 2, and 4, including forward and gradient checks.
\end{minipage}}
\endgroup
\caption{Representative live-steering cases. In (a), the supervisor redirects an unproductive strategy. In (b), the supervisor identifies an implementation error. In both cases, the worker adopts the correction and completes the task.}
\label{fig:live-steering-cases}
\end{figure}

%% file: sections/related.tex
\section{Related Work}
\label{sec:related}

\paragraph{Agent systems and self-evolving agents.}
Language-model agents choose actions, invoke tools, incorporate environment feedback, and retain useful experience across multiple steps \citep{toolformer,voyager}.
Existing systems improve execution through single-agent self-correction such as ReAct, Self-Refine, and CRITIC; delegated roles in AutoGen, MetaGPT, Magentic-One, and Claude Code \citep{autogen,metagpt}; or post-hoc memories derived from completed tasks in Reflexion and ExpeL.
A growing line of work instead evolves persistent agent components while keeping the base model fixed.
ADAS searches over agent programs, ACE evolves a context playbook \citep{adas,ace}, AutoHarness synthesizes code harnesses, and Meta-Harness and AHE optimize broader harness components from execution traces and evaluation feedback.
Group-Evolving Agents shares experience across evolving agent populations \citep{gea}; EvoSkill and Memento-Skills refine reusable skills, while Mem$^2$Evolve co-evolves experience and agent assets \citep{evoskill,memento,mem2evolve}.
Continual Harness adapts prompts, subagents, skills, and memory within a reset-free run, while Self-Harness mines weaknesses and retains validated harness edits \citep{continualharness, selfharness}.
Most of these methods generate and select an improved context, agent, or harness across completed work.
\textsc{PILOT} additionally acts on the current run: live steering redirects the active worker during execution, while live self-evolution records reusable harness knowledge during the same supervision process.

\paragraph{Long-horizon agent tasks.}
Agent benchmarks are moving from short, isolated tasks toward longer sequences of interdependent actions in stateful environments.
WebArena and OSWorld introduced realistic web and desktop interaction, while SWE-bench brought repository repair into executable environments \citep{webarena,osworld}.
Terminal-Bench~2.0 extends this direction to multi-step system and engineering work performed through an interactive shell.
Recent benchmarks make the horizon itself a central design axis.
LongCLI-Bench evaluates sequential engineering projects with step-level scoring and finds that state-of-the-art agents pass fewer than 20\% of tasks, with most runs stalling before 30\% completion \citep{longclibench}.
OSWorld~2.0 extends computer-use evaluation to workflows that take people a median of 1.6 hours and require hundreds of tool calls \citep{osworld2}, while EdgeBench studies ultra-long real-world tasks that sustain at least 12 hours of agent--environment interaction \citep{edgebench}.
This shift makes goal retention, interpretation of environment feedback, and recovery from compounding errors central evaluation concerns.
\textsc{PILOT} targets this runtime layer by closing the loop between recovery and accumulation: live steering acts on an ongoing trajectory, and live self-evolution makes supervision reusable across related long-horizon tasks.

%% file: sections/conclusion.tex
\section{Conclusion}

We presented \textsc{PILOT}, a supervisor--worker harness for live self-improvement of long-horizon agents.
\textsc{PILOT} implements live self-improvement through two mechanisms: live steering lets a separate supervisor redirect an active worker during execution, and live self-evolution turns procedures and failure modes observed during supervision into reusable harness knowledge.
Across two frozen backbones and three benchmarks, \textsc{PILOT} ranks first in five of six backbone--benchmark combinations.
In the self-improvement setting, \textsc{PILOT}'s best observed Terminal-Bench~2.0 pass rate increases by 14.6 points with GLM-5.1 and 12.4 points with Kimi-K2.6, while the reusable skill set grows and mean output tokens per evaluated task fall by 42.9\% and 47.4\%, respectively.
These results show that correction and accumulation are most useful when they form one continuous loop: supervision can recover the current trajectory, update the persistent harness, and improve the work of later agents.

\paragraph{Limitations.}
Iterative self-improvement repeats every task across many iterations, making additional benchmarks and backbones substantially more expensive than a single inference run.
This cost limits the current evaluation to three benchmarks and two open-weight backbones; broader coverage and proprietary models remain future work.
The supervisor and worker share the same backbone, reflecting common use but leaving heterogeneous pairings and their trade-offs among oversight quality, task performance, and cost unexplored.

\section*{Full Author List}
\begin{flushleft}
\small
\begingroup
\renewcommand{\thefootnote}{*}
\hypersetup{linkcolor=abyss}
Yang Xiao\footnote{Equal contribution.}, Yusong Sun\footnotemark[1], Haoyi Wu, Wenyang Hui, Wen Da, Zhaokai Luo, Mu Chuan, Yao Hu, Wenjie Li,\\
Chengyue Jiang
\endgroup
\end{flushleft}

%% file: sections/appendix.tex
\section{Appendix}

\subsection{Excluded SWE-bench tasks}
\label{app:excluded-swe}
\input{sections/appendix-excluded-swe}

\subsection{PILOT pseudocode}
\label{app:pilot-pseudocode}

Algorithm~\ref{alg:pilot} summarizes one supervisor session coordinating multiple worker sessions.
Workers may run concurrently or be spawned later in the same episode; every event and action is associated with its worker index $j$.
The five emphasized channel operations correspond directly to Figure~\ref{fig:framework}: workers emit \textbf{Notification}, \textbf{Question}, and \textbf{Result} events, while the supervisor can issue \textbf{Steer} and \textbf{Abort} actions.

\begin{algorithm}[H]
\caption{The coupled live steering and live self-evolution loops across worker sessions in \textsc{PILOT}.}
\label{alg:pilot}
\begin{minipage}[t]{0.49\linewidth}
\centering\textbf{Supervisor loop}\par\smallskip
\begin{algorithmic}
\Require task $\tau$, frozen model $\theta$, harness $H$ with skills $\mathcal{K}$ and memory $\mathcal{M}$
\Repeat
  \If{another worker session is needed}
    \State choose an objective $\tau_j$ for the worker
    \State $W_j \gets \textsc{Spawn}(\theta, \tau_j, H)$
  \EndIf
  \State receive an event $(j,e)$ from any active $W_j$
  \If{the event is a question}
    \State reply to $W_j$
  \EndIf
  \If{redirection of $W_j$ is warranted}
    \State inspect the relevant portion of $\xi_j$
    \State \textbf{Steer:} queue guidance for $W_j$
  \ElsIf{continuing $W_j$ is no longer useful}
    \State \textbf{Abort:} interrupt $W_j$
  \EndIf
  \If{reusable knowledge is found in $\xi_j$}
    \State $H' \gets \textsc{Update}(H, \xi_j)$
    \State $H \gets H'$
  \EndIf
  \If{$e$ is a result, error, or abort}
    \State mark $W_j$ as settled
  \EndIf
\Until{the task is resolved or no further worker is needed}
\State \Return outcome and updated harness $H$
\end{algorithmic}
\end{minipage}
\hfill
\begin{minipage}[t]{0.49\linewidth}
\centering\textbf{Worker-session loop (for each $W_j$)}\par\smallskip
\begin{algorithmic}
\Require objective $\tau_j$, frozen model $\theta$, harness $H$ at spawn time
\Repeat
  \State incorporate queued supervisor guidance
  \State reason, act in $\mathcal{E}$, and extend $\xi_j$
  \If{the worker decides to surface an update}
    \State \textbf{Notification:} report progress, an intermediate result, or a risk
    \State continue execution
  \EndIf
  \If{the worker decides supervisory input is needed}
    \State \textbf{Question:} send the question and pause
    \State receive the reply and resume
  \EndIf
\Until{completion, error, or abort}
\If{complete}
  \State \textbf{Result:} the runtime sends $(j,\text{result})$ to the supervisor
\EndIf
\end{algorithmic}
\end{minipage}
\end{algorithm}

\subsection{Self-improvement instruction}
\label{app:self-improvement-instruction}

The self-improvement setting prepends the same instruction to every task for every harness evaluated in that setting.
Only the harness-native skill and memory paths differ.
The instruction below uses \texttt{[SKILL\_PATH]} and \texttt{[MEMORY\_PATH]} for those substitutions.

\begin{center}
\small
\begin{tabular}{@{}lll@{}}
\toprule
Harness & \texttt{[SKILL\_PATH]} & \texttt{[MEMORY\_PATH]} \\
\midrule
PILOT & \nolinkurl{~/.pilot/skills} & \nolinkurl{~/.pi/agent/AGENTS.md} \\
Pi & \nolinkurl{~/.pi/agent/skills} & \nolinkurl{~/.pi/agent/AGENTS.md} \\
OpenCode & \nolinkurl{~/.config/opencode/skills} & \nolinkurl{~/.config/opencode/AGENTS.md} \\
\bottomrule
\end{tabular}
\end{center}

\begin{quote}
\scriptsize
You already have a library of reusable skills (auto-loaded from \texttt{[SKILL\_PATH]}) plus long-term notes (\texttt{[MEMORY\_PATH]}), distilled from prior experience. They're available right now---use them.

This library keeps growing: whatever new you save also persists, so the next run of a similar task can replay your approach instead of rediscovering it. So whenever you solve a task that took real work---finding the right tool or library, setting up an environment, working out a multi-step approach---capture that approach as a skill named after the task (e.g., ``\texttt{<task-name>}''), EVEN IF it felt routine in hindsight. The test is NOT ``was this non-obvious?''---it's ``if a similar task came up later, would replaying this save me from figuring it out again?'' If yes, save it. Record the exact tools, libraries, commands and setup you used to crack it. If solving this task also built on skills already in your library, note which ones in the new skill---so the reuse is tracked and a later run can combine them. Use EXACTLY ONE skill per task, named EXACTLY after the task name (a task ``\texttt{chess-best-move}'' $\rightarrow$ a skill ``\texttt{chess-best-move}'', not a descriptive name). Before writing a skill, check whether your library already has one for this task: if it exists, update it ONLY when this run genuinely found a better or more reliable approach (a fix, a sturdier step, a cleaner method)---if it's already good enough, leave it as-is, don't churn it; if it doesn't exist yet, create it. Never add a second skill for a task that already has one. Use \texttt{AGENTS.md} for the more general cross-task lessons. Err on the side of saving.

Here's the task:
\end{quote}

%% file: sections/appendix-excluded-swe.tex
We exclude 43 SWE-bench Multilingual tasks and 198 SWE-bench Pro tasks whose pinned environments cannot bootstrap the JavaScript agent runtimes required by the evaluated harnesses. The SWE-bench Pro exclusions also cover task images whose Node.js versions are below the runtimes' minimum requirement. We apply the same exclusions to every harness. For reproducibility, the complete excluded-task lists will be released in the \href{https://github.com/XiaoYang66/Pilot}{project GitHub repository}.